\documentclass[letterpaper, 10pt, conference]{ieeeconf}
\IEEEoverridecommandlockouts 
\pdfoutput=1
\usepackage{geometry}
\usepackage[export]{adjustbox}
\usepackage{amssymb}
\usepackage{bm}
\usepackage[caption=false]{subfig}
\usepackage{float} 
\usepackage{tabularx}
\usepackage{glossaries}
\usepackage[bookmarks=true]{hyperref}
\usepackage[noabbrev]{cleveref}      
\usepackage{multirow}
\usepackage[binary-units=true]{siunitx}

\usepackage[ruled,vlined]{algorithm2e}

\newcommand{\R}{\mathbb{R}}

\DeclareMathOperator*{\argmin}{arg\,min}

\Crefname{section}{Section}{Section}
\Crefname{subsection}{Section}{Section}
\Crefname{subsubsection}{Section}{Section}
\Crefname{figure}{Fig.}{Fig.}

\hypersetup{
    pdfauthor   = {Paarth Shah and Avadesh Meduri},%
    pdftitle    = {Rapid Convex Optimization of Centroidal Dynamics using Block Coordinate Descent},%
    pdfsubject  = {Robotics},%
    pdfkeywords = {Robots, Trajectory Optimization, Centroidal Dynamics, Block Coordinate Descent, Convex Relaxation, Direct Transcription}%
}

\title{\LARGE \bf
Rapid Convex Optimization of Centroidal Dynamics \\ using Block Coordinate Descent
}

\author{Paarth Shah$^{*1}$, Avadesh Meduri$^{*2}$, Wolfgang Merkt$^{1}$, Majid Khadiv$^{3}$, Ioannis Havoutis$^{1}$, Ludovic Righetti$^{2,3}$
\thanks{*These authors contributed equally}
\thanks{$^{1}$Oxford Robotics Institute, University of Oxford, England.
        {\tt\small {\{paarth,wolfgang,ioannis\}}@robots.ox.ac.uk}}%
\thanks{$^{2}$Tandon School of Engineering, New York University, Brooklyn, USA.
        {\tt\small {\{am9789,ludovic.righetti\}}@nyu.edu}}%
\thanks{$^{3}$Max Planck Institute for Intelligent Systems, T\"ubingen, Germany.
        {\tt\small
        mkhadiv@tuebingen.mpg.de    }}

\thanks{This work was supported in part by New York University, the European Union’s Horizon 2020 research and innovation program (grant agreement 780684), the National Science Foundation (grants 1825993, 1932187, 1925079 and 2026479), the UKRI/EPSRC with grants [EP/S002383/1], [EP/R026084/1] and [EP/R026173/1]. This work was part of the Human-Machine Collaboration Programme, supported by a gift from Amazon Web Services.}
\thanks{$^{1}$Paarth Shah was supported by an AWS Lighthouse Scholarship}
}

\begin{document}
\bstctlcite{IEEEexample:BSTcontrol}

\maketitle
\thispagestyle{empty}
\pagestyle{empty}

\begin{abstract}

In this paper we explore the use of block coordinate descent (BCD) to optimize the centroidal momentum dynamics for dynamically consistent multi-contact behaviors. The centroidal dynamics have recently received a large amount of attention in order to create physically realizable motions for robots with hands and feet while being computationally more tractable than full rigid body dynamics models. Our contribution lies in exploiting the structure of the dynamics in order to simplify the original non-convex problem into two convex subproblems. We iterate between these two subproblems for a set number of iterations or until a consensus is reached. We explore the properties of the proposed optimization method for the centroidal dynamics and verify in simulation that motions generated by our approach can be tracked by the quadruped Solo12. In addition, we compare our method to a recently proposed convexification using a sequence of convex relaxations as well as a more standard interior point method used in the off-the-shelf solver IPOPT to show that our approach finds similar, if not better, trajectories (in terms of cost), and is more than four times faster than both approaches. Finally, compared to previous approaches, we note its practicality due to the convex nature of each subproblem which allows our method to be used with any off-the-shelf quadratic programming solver.

\end{abstract}

\section{Introduction} \label{sec:introduction}
The use of optimization to generate movements for robots with hands and feet has been studied extensively over the past years. The problem is inherently complex due to the nonlinear nature of the dynamics, the non-convex cost landscape, and the requirement that computed trajectories must eventually be executable on real robots. 

In order to generate trajectories for online control, early research focused on planning using template models \cite{template_full}. These simplified models are low-dimensional approximations that capture the nature of the dynamics and frequently remove the nonlinearities and non-convexities which allows fast online re-computation due to their linear nature. The most widely studied simplified model in humanoid control has been the linear inverted pendulum and its many variations \cite{Kajita,herdt,englesberger,hopkins}. In these works, the linearity is exploited to efficiently solve for center of mass trajectories, footstep locations, or both, online. Although these methods have proven to be highly effective, they do not generalize to arbitrary terrains due to the inherent assumptions made in their formulations and focus on legged locomotion rather than the arbitrary multi-contact problem (e.g. using hands) which limits their versatility. 
\begin{figure}[t]
    \centering
    \includegraphics[width=\linewidth]{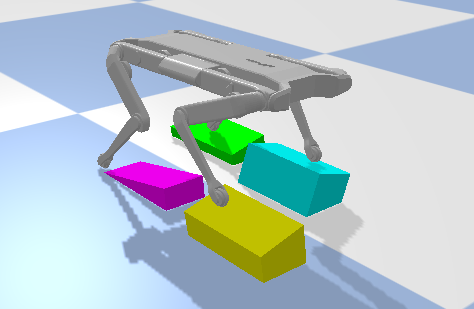}
    \caption{Simulation of a trajectory computed using our method on complex terrain with the quadruped Solo12.}
    \label{fig:intro_figure}
\end{figure}
%
Recently, there has been an increasing interest in developing full body motions for floating base robots using the centroidal dynamics model \cite{orin}. The centroidal dynamics are a reduced order representation of the full dynamics of the robot that considers the momentum wrench at the center of mass. One of the main benefits to this approach is the ability to utilize the full potential of the entire robot (e.g. arms and legs) to interact with arbitrary environments while also obeying the rigid body dynamics. So far, these methods have shown impressive results \cite{dai_fullbody, kino-dyn,carpentier, SCR}, with the latter two demonstrating capabilities on hardware itself.

The centroidal dynamics, however, are non-convex, which makes motion planning problems difficult to solve. The authors of \cite{dai} use a worst-case $\ell_{1}$ bound on the angular momentum in order to make the problem convex. \cite{carpentier} used an efficient multiple shooting approach, but to the best of our knowledge, this implementation is closed-source due to use of a proprietary solver, MUSCOD-II. In \cite{kino-dyn}, the non-convexity was dealt with by using a difference of quadratic functions which was further exploited and optimized in \cite{SCR}. In addition to providing the decomposition approach, \cite{kino-dyn, dai_fullbody} also proposed a method for creating full-body motions using an iterative approach of alternating the optimization of the centroidal dynamics and whole-body kinematics. Although the results from \cite{SCR} were impressive, their framework needs to solve second-order cone programs which are more computationally demanding than solving simple quadratic programs (QP). Specifically, they used a customized variant of the ECOS solver \cite{ecos}, an interior-point solver which is therefore difficult to warm-start for model-predictive control applications.

In this paper, we propose a block coordinate descent (BCD) approach to solve the centroidal dynamics trajectory optimization problem \cite{multi_convex_boyd}. The main idea of the approach is to leverage the inherent structure of the problem in order to simplify the non-convexity into two simpler, convex subproblems. By utilizing the sparse structure of the multi-contact locomotion problem, we show that we are able to efficiently generate and track different types of whole-body motions including challenging maneuvers that require tight tracking of angular momentum.

Unlike previous methods which rely on complicated decomposition procedures and customized solvers (both of which are difficult to implement), our approach can instead be easily implemented using any off-the-shelf QP solver. 

The proposed method also has guaranteed convergence to a feasible solution under the assumption that the problem is well-posed unlike methods that rely on off-the-shelf nonlinear solvers such as IPOPT \cite{ipopt}. Although the solutions may not converge to a local minima, we are guaranteed to converge to feasible motions at every iteration \cite{bcd}.
We argue that feasibility, in terms of constraint satisfaction, is more important than optimality, as the local minima found by off-the-shelf solvers are often arbitrary. We show that in practice, our algorithm converges quickly and finds high quality trajectories that can be tracked in simulation.

Finally, we find that our method is often multiple times faster than the state of the art. Due to the structure of our optimal control problem as well as the convexity of each subproblem, we are able to leverage the maturity of QP solvers that offer features that can inherently exploit the sparsity patterns in our problem. 




\section{Whole Body Trajectory Optimization} \label{sec:problem_statement}
The equations of motion for a floating-based rigid body dynamics robot can be written as
\begin{equation}
    \bm{M}(\bm{q})\ddot{\bm{q}}+\bm{h}(\bm{q},\dot{\bm{q}}) = \bm{S}^{T}\bm\tau_{j} + \sum \limits_{i=0}^{n_{c}}\bm{J}_{i}^{T} \bm\lambda_{i}
\end{equation}
where $\bm{q} = [\bm{x}^{T}~\bm{q}_{j}^{T}]^{T}$ describes the robot configuration and comprises both the floating base position and orientation expressed with respect to a fixed inertial frame, $\bm{x} \in \mathbb{SE}(3)$, and joint positions, $\bm{q}_{j}$, of the robot. $\bm{M}(\bm{q}) \in \R^{(n_{j}+6) \times (n_{j}+6)}$ is the mass inertia matrix, $\bm{h}(\bm{q}, \bm{\dot{q}}) \in \R^{(n_{j}+6)}$ contains the Coriolis, centrifugal, gravity, and friction forces, $\bm{S} = [\bm{0}^{n_{j} \times 6}~\bm{I}^{n_{j}\times n_{j}}]$ is the actuator selection matrix that defines the underactuation of the robot, $\bm{\tau}_{j} \in \R^{n_{j}}$ is the vector of joint torques, $\bm{J}_{i}$ are the end-effector jacobians and $\bm{\lambda}_{i}$ are the forces due to external contacts acting on the robot.

For underactuated robots (i.e. robots with more degrees of freedom than number of controllable joints), we can decompose our dynamics into the actuated (subscript a) and unactuated (subscript u) dynamics as follows 
\begin{subequations}
\begin{align}
    \bm{M_{a}}&(\bm{q})\ddot{\bm{q}}+\bm{h}_{a}(\bm{q},\dot{\bm{q}}) = \bm\tau_{j} + \sum \limits_{i=0}^{n_{c}}\bm{J}_{i,a}^{T} \bm\lambda_{i} \label{decompose_newton_euler_a}\\
   \bm{M_{u}}&(\bm{q})\ddot{\bm{q}}+\bm{h}_{u}(\bm{q},\dot{\bm{q}}) = \sum \limits_{i=0}^{n_{c}}\bm{J}_{i,u}^{T} \bm\lambda_{i} \label{decompose_newton_euler_b}
\end{align}
\label{decompose_newton_euler}
\end{subequations}

Equation \eqref{decompose_newton_euler_b} describes the change in momentum of the robot given external forces, $\lambda$. As previously described in \cite{kino-dyn}, the actuated part of the dynamics provides the necessary torques to achieve combinations of the desired accelerations, $\bm{\ddot{q}}$, and contact forces, $\lambda$. Under the assumption of enough actuation torque, this allows us to ignore the actuated part of the dynamics, and focus solely on creating motions for the underactuated floating base using purely the external forces and torques. The underactuated dynamics are equivalent to the centroidal dynamics of the robot \cite{wieber2006holonomy} and when expressed at the robot center of mass (CoM) can be written as
\begin{equation}
    \bm{\dot{h}} = 
    \begin{bmatrix}
    \bm{\dot{r}} \\
    \bm{\dot{l}} \\
    \bm{\dot{k}}
    \end{bmatrix} = 
    \begin{bmatrix}
    \frac{1}{m}\bm{l} \\
    m\bm{g} + \sum_{e} \bm{f}_{i} \\
    \sum_{e}(\bm{p}_{e} + \bm{R}_{e,t}^{x,y}\bm{z}_{e} - \bm{r}) \times \bm{f}_{e} + \bm{\tau}_{e}
    \end{bmatrix}.
    \label{centroidal_ode}
\end{equation}
Here, $\bm{r}$ is the center of mass location, $\bm{l}$ is the linear momentum of the center of mass, $\bm{k}$ is the angular momentum of the center of mass, $\bm{f}_{e}$ are the external forces on the robot, $\bm{p}_{e}$ are the robot end-effector locations in the inertial frame, $\bm{z}_{e}$ the centers of pressure for each contact, $\bm{R}_{e,t}^{x,y} \in \R^{3 \times 2}$ are the first two columns of the rotation matrix $\bm{R}_{e,t}$ which rotates maps quantities from end-effector frame to inertial frame, $\bm{\tau}_{e}$ the torques at each center of pressure (e.g. torques induced by flat feet of a robot leg), $m$, is the robot mass, and $\bm{g}$ is the gravity vector.

Using this form, \cite{kino-dyn} and \cite{dai_fullbody} suggested a decomposition to create whole-body motions. \cite{kino-dyn} proposed an alternating method for the kinematics and dynamics of the robot by finding dynamically feasible trajectories using the centroidal dynamics then solving an inverse kinematics problem for the whole body of the robot. The output of this alternating process are whole-body motions that can then be tracked by a controller such as in \cite{momentum_id}.

\subsection{Trajectory Optimization of Centroidal Dynamics}
In this paper, we focus on the optimization of the centroidal dynamics, Eq. \eqref{centroidal_ode}. Specifically, we are looking to find dynamically feasible trajectories, i.e. motions that optimize for the end-effector forces and torques subject to the non-convex constraints of the centroidal momentum. We assume that a contact surface is given and that the timing of each contact is fixed. Contact sequences can be found using a contact planner \cite{deits,sl1m,lin2018humanoid}. The optimization problem we are trying to solve can be written as follows
\begin{subequations}
\begin{gather}
    \underset{\bm{h}, \bm{p}_{e}, \bm{f}_{e},\bm{\tau}_{e}, \bm{z}}{\text{min}} \quad 
    \sum \limits_{t=0}^{N}  \Psi_{t}(\bm{h, p_{e}, z_{e}, f_{e}, \bm{\tau_{e}})}\ +\ \phi_{t}(\bm{h}_{t} - \bm{h}_{t}^{kin})  \label{traj_opt_cost}\\
    \text{s.t. } \ 
    \bm{h}_{t} = 
    \begin{bmatrix}
    \bm{r}_{t} \\
    \bm{l}_{t} \\
    \bm{k}_{t}
    \end{bmatrix} = 
    \begin{bmatrix}
    \bm{r}_{t-1} + \frac{1}{m}\bm{l}\Delta t \\
    \bm{l}_{t-1} + m\bm{g}\Delta t + \sum_{e} \bm{f_{e,t}} \Delta t \\
    \bm{k}_{t-1} + \sum_{e} \bm{\kappa}_{e,t} \Delta t
    \end{bmatrix} \\
    \bm{\kappa}_{e,t} = (\bm{p}_{e,t} - \bm{r_{t}}) \times \bm{f}_{e,t} + \bm{\gamma}_{e,t} \label{eqn:angular_1} \\
    \bm{\gamma}_{e,t} = (\bm{R}_{e,t}^{x,y} \bm{z}_{e,t}) \times \bm{f}_{e,t} + \bm{\tau}_{e,t} \label{eqn:angular_2} \\
    \bm{p_{e,t}} \in \mathcal{U}(\mathcal{S}) \\
    \bm{z}_{e,t}^{x,y} \in [{}^{min}\bm{z}^{x,y}, {}^{max}\bm{z}^{x,y}] \\
    \label{eqn:frict-cone}
    |\bm{f}_{e,t}^{x}| \leq \ \bm{R}_{e,t}\bm{f}_{e,t}^{z}, \ |\bm{f}_{e,t}^{y}| \leq \ \bm{R}_{e,t}\bm{f}_{e,t}^{z}, \ \bm{R}_{e,t}\bm{f}_{e,t}^{z} \geq 0 \\ 
    \label{eqn:kin-limit}
    \parallel \bm{p}_{e,t} - \bm{r}_{t} \parallel \leq \mathcal{L}^{max}
\end{gather}
\label{eqn:full_traj_opt}
\end{subequations}
where constraints \eqref{eqn:frict-cone} are the pyramidal friction cone constraints, \eqref{eqn:kin-limit} is an $\ell_{1}$-norm approximation of the kinematic limit of the end-effectors,  and \eqref{traj_opt_cost} minimizes a quadratic sum of the running cost on the discretized dynamics of the state $\Psi$ and cost of tracking the output of the kinematic optimization, $\bm{\phi_{t}}$. We note that while some centroidal optimization approaches \cite{SCR} uses second-order cones ($\ell_{2}$-norms) for both the kinematic limit and friction cone constraints, we use linear approximations for both. 
Solving this problem efficiently is in general hard due to the cross product in \eqref{eqn:angular_1} and \eqref{eqn:angular_2} which introduce non-convex constraints.

\section{Block Coordinate Descent} \label{sec:bcd}
In this section, we give a brief overview of the block coordinate descent method used in the subsequent sections of the paper. We first detail the main idea and general approach, and later discuss the convergence properties of the chosen formulation and update methodology. 

We are interested in optimization problems of the form
\begin{equation}
\begin{aligned}
    \underset{\bm{x} \in X}{\text{min}} \quad  F(\bm{x}_{1},\cdots,\bm{x}_{s}) + \sum \limits_{i=1}^{s}r_{i}(\bm{x}_{i})
\end{aligned}
\end{equation}
where the variable $\bm{x}$ is decomposed into $\bm{s}$ blocks, the set $X$ is closed and a block multi-convex subset of $\R^{n}$. Note that the set $X$ may be non-convex over $\bm{x} = (\bm{x}_{1},\cdots,\bm{x}_s)$. $r_{i}$ are extended value functions which mean $r_{i}(\bm{x}_{i}) = \infty$ if $\bm{x}_{i}$ $\notin$ dom($r_{i}$) and can be used as indicator functions for convex sets.

We call a set $X$ block multi-convex if each block of variables is convex, that is, for each $i$ and fixed $(s-1)$ blocks $\bm{x}_{1},\cdots,\bm{x}_{i-1},\cdots,\bm{x}_{s}$ the set 
\begin{equation}
\begin{aligned}
    & X_{i}(\bm{x}_{1}...\bm{x}_{i-1}, \bm{x}_{i+1}, ...\bm{x}_{s}) \triangleq \\
    & \{ {\bm{x}_{i} \in \R^{n}} : (\bm{x}_{1}, ..., \bm{x}_{i-1}, \bm{x}_{i}, \bm{x}_{i+1}, ... \bm{x}_{s}) \in X \}
\end{aligned}
\end{equation}
is convex. We can then see, when all blocks except one are fixed, the function, $F$ is convex.

Block coordinate descent (BCD) of the Gauss-Seidel type minimizes $F$ cyclically over each of the individual blocks $x_{i}$ while fixing the other blocks to their latest updated values \cite{multi_convex_boyd, gauss-seidel}.
\begin{equation}
    \bm{x}_{i}^{k+1} = \underset{\bm{x}_{i}}{\argmin}\ {f}_{i}^{k+1}(\bm{x}_{i}^{k+1}) + r_{i}(\bm{x}_{i})
\end{equation}
The general block coordinate descent method for non-convex problems, however, has no guarantees of convergence (either to a local minimum or otherwise) and may cycle infinitely. In order to address this, we use a proximal update when updating and solving each block
\begin{equation}
    \bm{x}_{i}^{k+1} = \underset{\bm{x}_{i}}{\argmin}\ {f}_{i}^{k+1}(\bm{x}_{i}^{k+1}) + \frac{L_{i}^{k}}{2}||\bm{x}_{i}^{k+1} - \bm{x}_{i}^{k}||^{2} + r_{i}(\bm{x}_{i})
\end{equation}
where $L_{i}^{k}$ is a non-zero regularization parameter and $|| \cdot ||$ is the $\ell_2$-norm. The proximal parameter, $L_{i}^{k}$, is in practice used to regularize the current solution to ensure we do not stray too far from the previous solution (i.e. introduces damping) and may change at every iterate. Using this method, we are guaranteed to converge to a feasible solution \cite{bcd}. 

An important note is that that this does not necessarily mean we are guaranteed to converge to a local minimum of the original optimization problem $F$. Rather, we are only guaranteed a feasible trajectory (i.e. all constraints are satisfied). We will see in the following subsections how an appropriate choice of blocks and use of the proximal update parameter allows us to converge to reasonable solutions for the centroidal dynamics optimization.

\subsection{Block Coordinate Descent for Centroidal Dynamics}
In order to solve the centroidal optimization problem, \autoref{eqn:full_traj_opt}, using the block coordinate descent method we first apply a change of variables similar to one introduced in \cite{SCR}. Specifically, we apply a change of variables to the cross products between the force, $\bm{f}$ and contact location ($\bm{p}_{e}-\bm{r}$) as well as the $\bm{f}$ and rotated ZMP in Eqs. \eqref{eqn:angular_1} and \eqref{eqn:angular_2} to combine these into one variable, $\bm{\ell}$:
\begin{equation}
\begin{split}
    \bm{\kappa}_{e,t} & = (\bm{p}_{e,t} - \bm{r_{t}} + \bm{R}_{e,t}^{x,y} \bm{z}_{e,t}) \times \bm{f}_{e,t} + \bm{\tau}_{e,t} \\
    & = \bm{\ell} \times \bm{f}_{e,t} + \bm{\tau}_{e,t} \\
    & = \begin{bmatrix}
        \phantom{-}0 & -\bm{\ell}_{e,t}^{z} & \phantom{-}\bm{\ell}_{e,t}^{y}\\
        \phantom{-}\bm{\ell}_{e,t}^{z} & \phantom{-}0 & -\bm{\ell}_{e,t}^{x} \\
        -\bm{\ell}_{e,t}^{y} & \phantom{-}\bm{\ell}_{e,t}^{x} & 0
        \end{bmatrix}
        \begin{bmatrix}
        \bm{f}_{e,t}^{x} \\
        \bm{f}_{e,t}^{y} \\
        \bm{f}_{e,t}^{y}
        \end{bmatrix}
        + \bm{\tau}_{e,t}
\end{split}
\end{equation}

Using this change of variables, we see that the non-convexity is in fact biconvex. Specifically, for a fixed set of $\bm{\ell}$, the problem is convex with respect to our forces, $\textbf{f}$, and vice-versa. This new change of variables then suggests the use of two-block minimizations. 

Our algorithm is outlined as follows: We first fix $\bm{\ell_{i}}$, and solve for forces $\bm{f_{i}}$ in one quadratic program which we will call the Force-QP, $\bm{\zeta}$. We then use the forces to solve for $\bm{\ell}$ in a second QP which we call the Contact-QP, $\bm{\nu}$. We iterate on this process until a consensus is found or a maximum number of iterations is reached after which one final Force-QP is run to generate fully dynamically consistent profiles (i.e. forces for the appropriate CoM, momentum, and end-effector location profiles). \autoref{Algorithm 1} provides an outline of the block coordinate descent algorithm.

\subsection{Force Quadratic Program}
The Force-QP solves the full centroidal dynamics problem, \autoref{eqn:full_traj_opt}, for a fixed $\bm{\ell}$. During each iteration, $k$, we increase the parameter $L_{i}$ by a factor $\alpha$. Due to the use of quadratic costs, the parameter $L_{i}$ in practice is used to regularize the solution from the previous solution. The Force-QP can be stated as follows
\begin{subequations}
\begin{gather}
    \begin{split}
        \underset{\bm{h}^{k}, \bm{f}^{k}_{e},\bm{\tau}^{k}_{e}, \bm{z}^{k}}{\text{min}} \quad 
        \sum \limits_{t=0}^{N} \bm{[} \Psi_{t}(\bm{h}^{k}, \bm{z}_{e}^{k}, \bm{f}^{k}_{e}, \bm{\tau_{e}}^{k})\ + \\ 
        \phi_{t}(\bm{h}_{t}^{k} - \bm{h}_{t}^{kin}) + L^{k, \zeta}(\bm{h}_{t}^{k, \zeta} - \bm{h}_{t}^{k-1, \nu}) \bm{] } \\
    \end{split}
    \\
    \text{s.t.} \ \  \bm{h}_{t}^{k} = 
    \begin{bmatrix}
        \bm{r}_{t}^{k} \\
        \bm{l}_{t}^{k} \\
        \bm{k}_{t}^{k}
    \end{bmatrix} = 
    \begin{bmatrix}
        \bm{r}_{t-1}^{k} + \frac{1}{m}\bm{l}_{t}^{k}\Delta t \\
        \bm{l}_{t-1}^{k} + m\bm{g}\Delta t + \sum_{e} \bm{f}^{k}_{e,t} \Delta t \\
        \bm{k}_{t-1}^{k} + \sum_{e} \bm{\kappa}_{e,t}^{k} \Delta t
    \end{bmatrix} \\
    \bm{\kappa}_{e,t}^{k} = (\bm{\ell}^{k-1, \nu}) \times \bm{f}_{e,t}^{k} + \bm{\tau}_{e,t} \label{eqn:force_qp_torque} \\
    |\bm{f}_{e,t}^{x}| \leq \ \bm{R}_{e,t}\bm{f}_{e,t}^{z}, \ |\bm{f}_{e,t}^{y}| \leq \ \bm{R}_{e,t}\bm{f}_{e,t}^{z}, \ \bm{R}_{e,t}\bm{f}_{e,t}^{z} \geq 0 \\
    \parallel \bm{p}_{e,t}^{k-1,\nu} - \bm{r}_{t}^{k,\zeta} \parallel \leq \mathcal{L}^{max}
\end{gather}
\end{subequations}

We note that the optimization problem does not regularize the momentum dynamics from the previous Force-QP but rather from the previous Contact-QP (in the case of the first iteration when no Contact-QP has been run, we do not regularize the center of mass location at all). 

\subsection{Contact Quadratic Program}
The Contact-QP is then used to solve for the length of the CoM wrench, $\bm{\ell}$, given the forces solved in the previous QP. Rather than optimizing over $\bm{\ell}$ directly, we need to remember the physics of the problem we are trying to solve; specifically that $\bm{\ell}_{t} = \bm{p}_{e,t} - \bm{r}_{t} + \bm{R}_{e,t}^{x,y} \bm{z}_{e,t}$. In order to give us finer control of individually tracking the end-effector location, $\bm{p}_{e,t}$, and center of mass, $\bm{r}_{t}$, we separate these individually in our QP. This gives us the following optimization problem:
\begin{subequations}
\begin{gather}
    \begin{split}
    \underset{\bm{r}^{k}, \bm{p}^{k}_{e}, \bm{l}^{k}, \bm{z}_{e}}{\text{min}} \quad 
    \sum \limits_{t=0}^{N} \bm{[} \Psi_{t}(\bm{r}^{k}, \bm{p}_{e}^{k}, \bm{l}^{k}, \bm{z}^{k})\\ + L^{k, \nu}(\bm{h}_{t}^{k, \nu} - \bm{h}_{t}^{k-1, \zeta}, \bm{p}^{k, \nu}_{e} - \bm{p}^{k-1, \nu}_{e}) \bm{] } \\
    \end{split}
    \\
    \text{s.t.} \ \  
    \begin{bmatrix}
    \bm{r}_{t}^{k} \\
    \bm{k}_{t}^{k}
    \end{bmatrix} = 
    \begin{bmatrix}
    \bm{r}_{t-1}^{k} + \frac{1}{m}\bm{l}_{t}^{k, \zeta}\Delta t \\
    \bm{k}_{t-1}^{k} + \sum_{e} \bm{\kappa}_{e,t}^{k} \Delta t
    \end{bmatrix} \ \\
    \bm{\kappa}_{e,t}^{k} = \bm{f}_{e,t}^{k} \times (\bm{r}_{t}^{k,\nu} - \bm{p}_{e}^{k,\nu}) + \bm{\gamma}_{e,t}^{k} \label{eqn:contact-qp_ang1}\\
    \bm{\gamma}_{e,t}^{k} = - \bm{f}_{e,t}^{k} \times (\bm{R}_{e,t}^{x,y} \bm{z}_{e,t}^{k}) \label{eqn:contact-qp_ang2} \\
    \bm{z}_{e,t}^{x,y} \in [{}^{min}\bm{z}^{x,y}, {}^{max}\bm{z}^{x,y}] \label{eqn:contact-qp_zmp} \\
    \bm{p}_{e,t}^{\nu} \in \mathcal{U}(\mathcal{S}) \\
    \parallel \bm{p}_{e,t}^{k,\nu} - \bm{r}_{t}^{k,\nu} \parallel \leq \mathcal{L}^{max}
\end{gather}
\end{subequations}
\label{eqn:contact-qp}
where we rearrange Eqs. \eqref{eqn:contact-qp_ang1}, \eqref{eqn:contact-qp_ang2} using the cross product identity
\begin{equation}
    \bm{a} \times \bm{b} = - \bm{b} \times \bm{a}
\end{equation}

Once again, we use $L_{i}$ to regularize the solution from the previous QP. However, because the momentum dynamics $\bm{h_{t}}$ appears in the Force-QP itself, we instead regularize the trajectories with the values from the previous Force-QP rather than the previous Contact-QP. As in the case of the Force-QP, we increase the value of $L_{i}$ during each iteration. We would also like to note the lack of state transition constraints for the linear momentum, $\bm{l}_{i}$. If we were to add these constraints into our formulation, this would prevent the center of mass from being able to alter and track the angular momentum and would only rely on the contact location $\bm{p}_{e}$ and the ZMP, $\bm{z}$. By relaxing these constraints, we instead regularize the linear momentum from the previous Force-QP which allows our optimization the freedom to use the center of mass to reduce momentum. We once again note, that since we use one final Force-QP before finishing the alternating process, CoM trajectories that may be invalid due to the removal of the linear momentum constraints are pushed within the constraint set to satisfy the dynamic criteria. 

\subsection{Convergence Criteria}
The convergence of the algorithm is dependent on the weights chosen, in particular, the weighting factor, $L_i$ and the scaling factor between iterations, $\alpha$. In practice, we find that a good stopping point is when the angular momentum profiles from one iteration to the next fall below some consensus threshold, $\epsilon_{f}$ or after $K$ iterations. $\epsilon_{f}$ is defined as follows:
\begin{equation}
    \epsilon_{f} = \dfrac{||\bm{\ell}^{k} - \bm{\ell}^{k-1}||^{2}}{N}
\end{equation}
where $N$ is the horizon of our optimal control problem.

\begin{algorithm}
    \SetKwInOut{Input}{Input}
    \SetKwInOut{Output}{Output}
    Initialize optimization variables: \bm{$f_{0}$}, \bm{$h_{0}$}, \bm{$\ell_{0}$}
    \\
    set k = 0 
    \\
    \While{$k <\ $ {maximum iterations} } {
      {
        \bm{$f^{k+1,\zeta}, h^{k+1, \zeta}$} = $\mathrm{QP_{Force}}$(\bm{$h^{k, \nu}$}, \bm{$\ell^{k, \nu}$})\\
        \bm{$L^{k+1, \zeta}$} = $\alpha$\bm{$L^{k, \zeta}$} \\
        \bm{$\ell^{k+1, \nu}$} = $\mathrm{QP_{Contact}}$(\bm{$h^{k+1, \zeta}$}, \bm{$f^{k+1, \zeta}$})\\
        \bm{$L^{k+1, \nu}$} = $\alpha$\bm{$L^{k,\nu}$} \\
        \If{$||\ell^{k} - \ell^{k-1}||^{2} / N \leq \epsilon_{f}$}{
            terminate
        }
        {}
      }
     }
     \bm{$f^{k+1,\zeta}, h^{k+1, \zeta}$} = $\mathrm{QP_{Force}}$(\bm{$h^{k, \nu}$}, \bm{$\ell^{k, \nu}$})\\
    \caption{Block Coordinate Descent for Biconvex Optimization}
    \label{Algorithm 1}
\end{algorithm}



\section{Experimental Setup} \label{sec:experiment_setup}

\subsection{Multi-contact control pipeline and platform}
In order to verify and test the profiles generated by the dynamics optimization, we utilize the open-source kino-dynamic trajectory optimization package \cite{kino-dyn-github}. Contact sequences can either be computed using the MIQCQP \cite{SCR} or set manually and are then given to our block coordinate descent framework. After our method finds a dynamically feasible motion, these are then sent to the kinematics optimizer. The resulting output of the framework is a whole-body motion which is then tracked by the whole-body controller (WBC) outlined in \cite{solo}. The controller uses feedback on the centroidal momentum combined with a desired task-space impedance plus a joint space PD controller to solve a QP for end-effector forces. These forces are then mapped to actuator torques using the jacobian transpose and executed and tracked as torque commands.

The motions were generated for a 12 degree of freedom quadruped, Solo12, which is simulated using the PyBullet simulation software \cite{pybullet}. 
Due to the use of a quadruped with point contacts, we eliminate the end-effector torque constraints from \autoref{eqn:force_qp_torque} as well as the ZMP constraints in \autoref{eqn:contact-qp_zmp} and \autoref{eqn:contact-qp_ang2} when generating motions.

\subsection{Solver details}
The proposed method was implemented in Python using the open source QP solver OSQP \cite{osqp}. 
All experiments were run with the solver settings shown in \autoref{table:1} located in the Appendix. 
All computations were performed using a single thread with an Intel Core i7-9850H CPU @ \SI{4.6}{\giga\hertz} and \SI{16}{\giga\byte} \SI{2666}{\mega\hertz} RAM.



\section{Results} \label{sec:results}
We tested our algorithm in several different scenarios from flat ground motion generation to multi-contact navigation on uneven terrain as well as more dynamic motions such as jumping and bounding. We would like to note that most trajectories were optimized over relatively long horizons (around \SI{5}{\second}) with several contact switches. Despite this, we were able to successfully track these open-loop using only the WBC without re-optimization of the trajectory (i.e. without model predictive control). This suggests that the computed trajectories are of good quality. For each motion, we either give a convex region for the contact optimization or fix the contact location and only allow the center of mass to reduce the angular momentum.

\subsection{Generating various locomotion behaviors}
We first generated and tracked motions for different types of gaits such as walking, trotting, and bounding. For simple motions, such as walking and trotting, the algorithm converges within two iterations (i.e. after $k$ = 2). For bounding, which requires finer control of angular momentum, a third iteration was required before the algorithm converged.

\begin{figure*}
    \subfloat[Momentum profiles and tracking of non-gaited motion]{%
        \begin{tabular}[b]{@{}c@{}}
            \includegraphics[width=0.49\linewidth, trim={0, 240, 0, 0},clip]{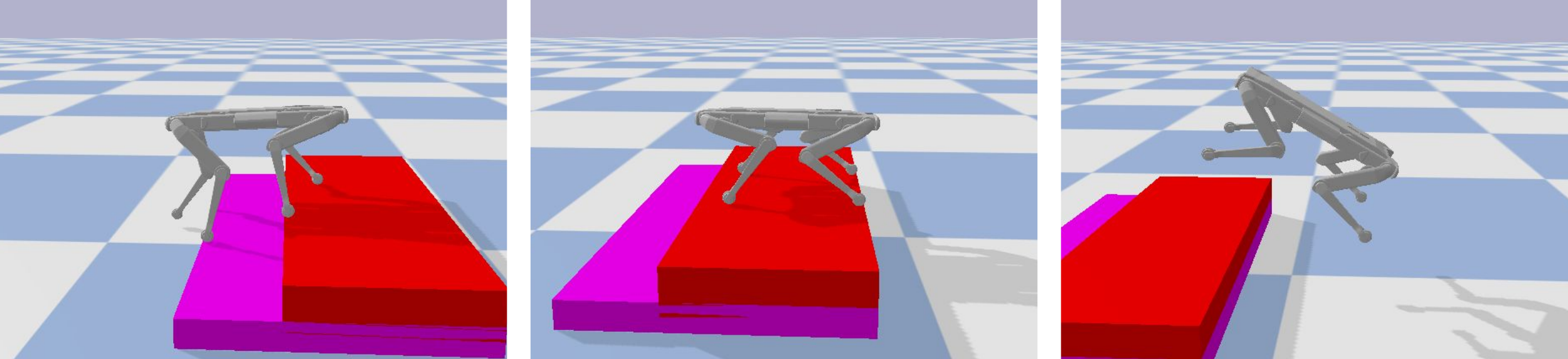}
            \\
            \includegraphics[width=.49\linewidth]{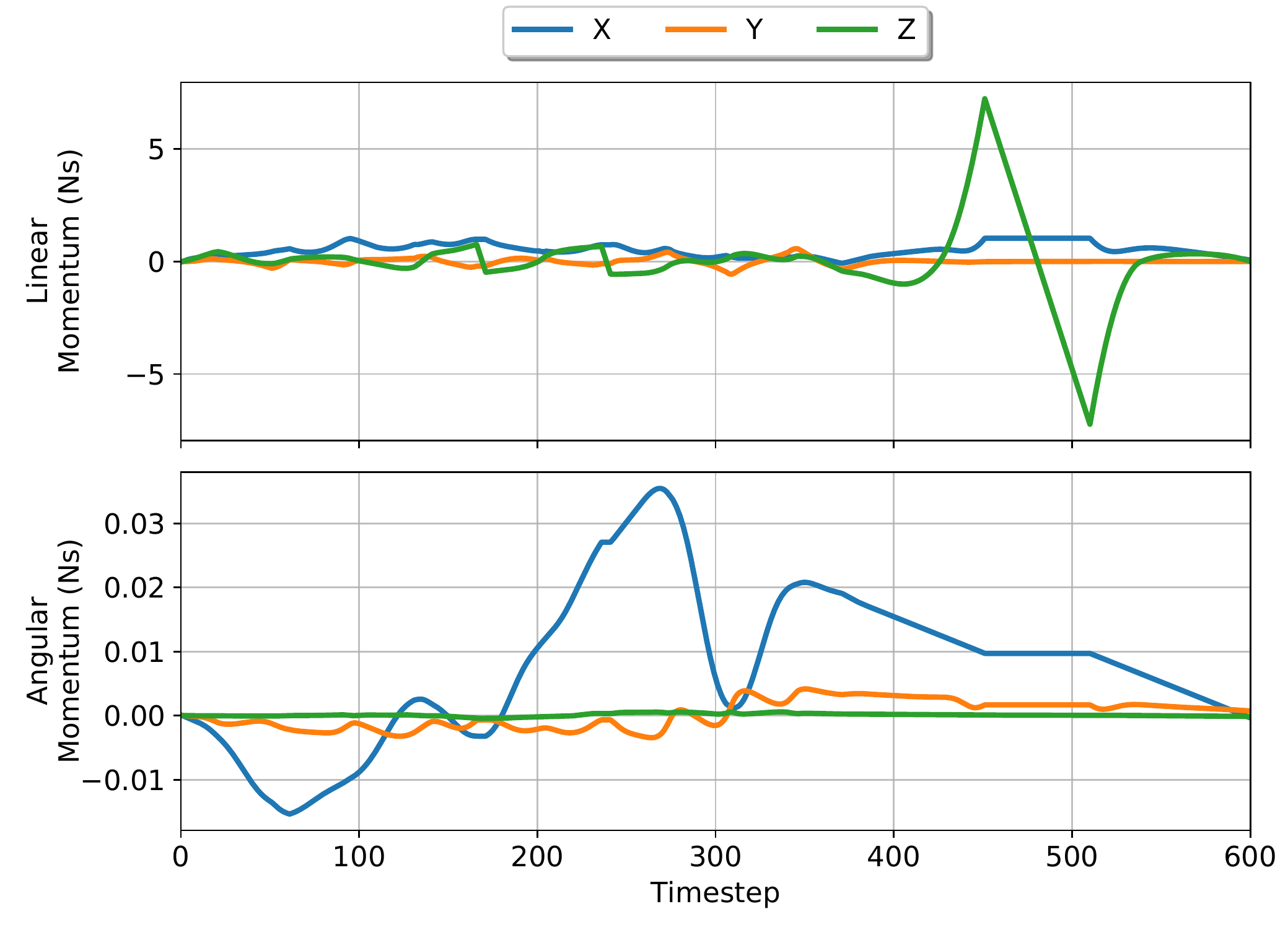}
            \label{fig:multicontact_traj}
        \end{tabular}
    }
    \hfill
    \subfloat[Momentum profiles and tracking of a bounding gait]{%
        \begin{tabular}[b]{@{}c@{}}
            \includegraphics[width=.49\linewidth, trim={0, 240, 0, 0},clip]{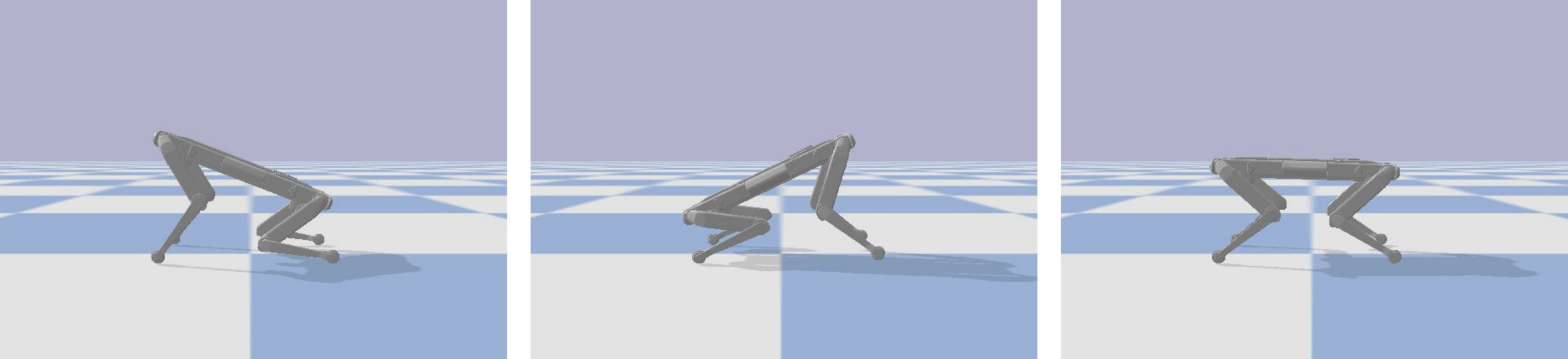}
            \\
            \includegraphics[width=.49\linewidth]{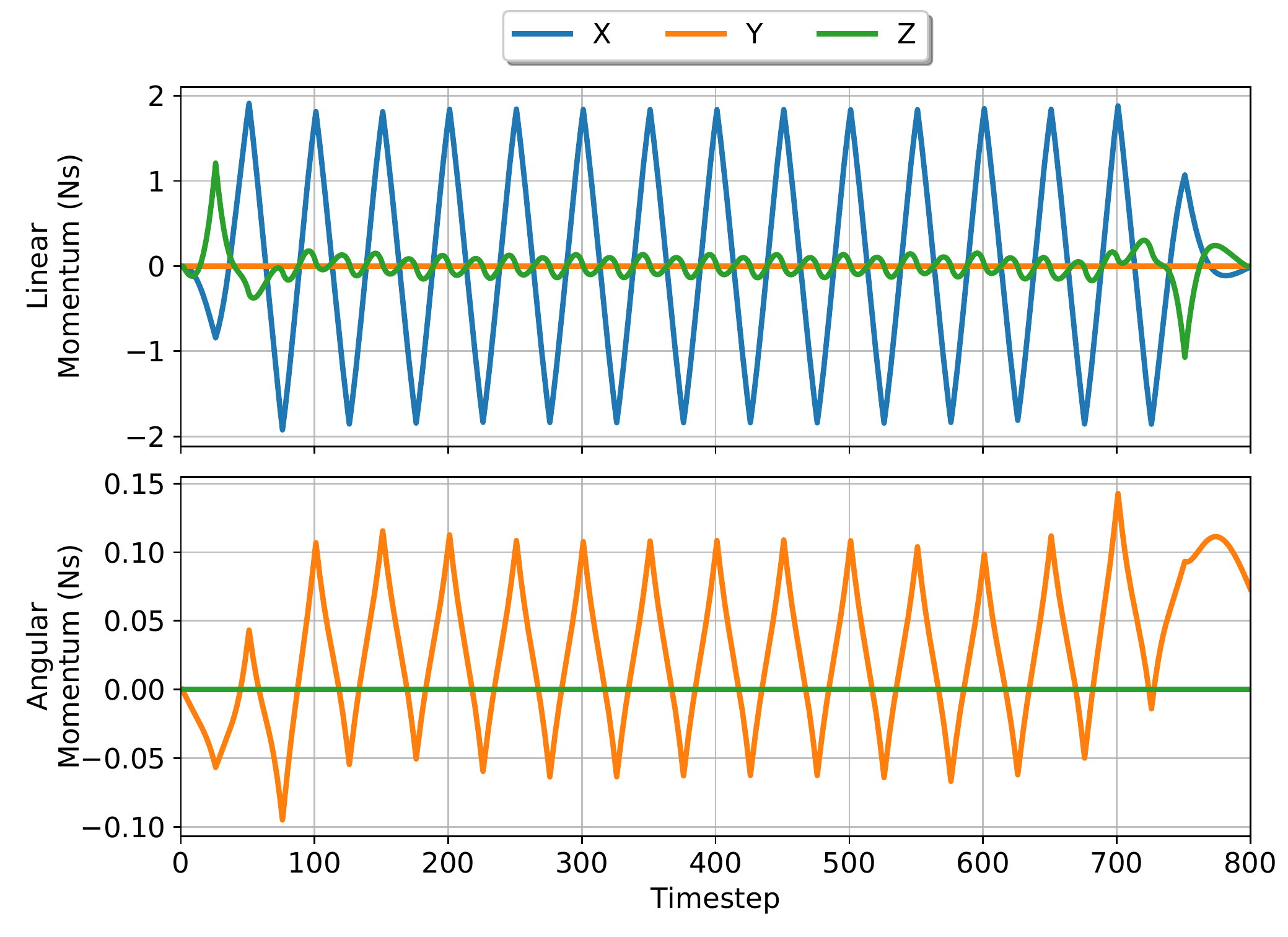}
            \label{fig:bounding}
        \end{tabular}
    }
    \caption{In this figure we plot the resulting profiles for two different multi-contact scenarios with multiple contact switches as well simulation screenshots showing the successful tracking of these motions. \Cref{fig:multicontact_traj} shows the profiles for a motion that included navigating over uneven terrain with a jump onto the ground which showcases the ability of our algorithm to compute trajectories with arbitrary, non-gaited contact switches together with highly dynamic motions. \Cref{fig:bounding}, shows the profiles for a bounding motion computed using our method. We set the contact locations to switch between the front and hind legs every \SI{0.25}{\second}. As we can see in both scenarios, we are able to generate high enough quality motions such that despite the numerous contact switches, we can successfully track each motion without the explicit re-optimization of the trajectories.}
    \label{fig:multicontact_motions}
    \vspace{-8pt}
\end{figure*}

Bounding in particular tests our algorithm's ability to generate and track trajectories that incur a high amount of angular momentum. Without explicitly re-optimizing angular momentum via model predictive control or receding horizon control, tracking of these types of trajectories is often difficult. In \Cref{fig:bounding} we show that despite this, we were able to generate and track such a motion for an 8 second horizon using only the WBC. 

\begin{figure}
    \center
    \includegraphics[width=\linewidth]{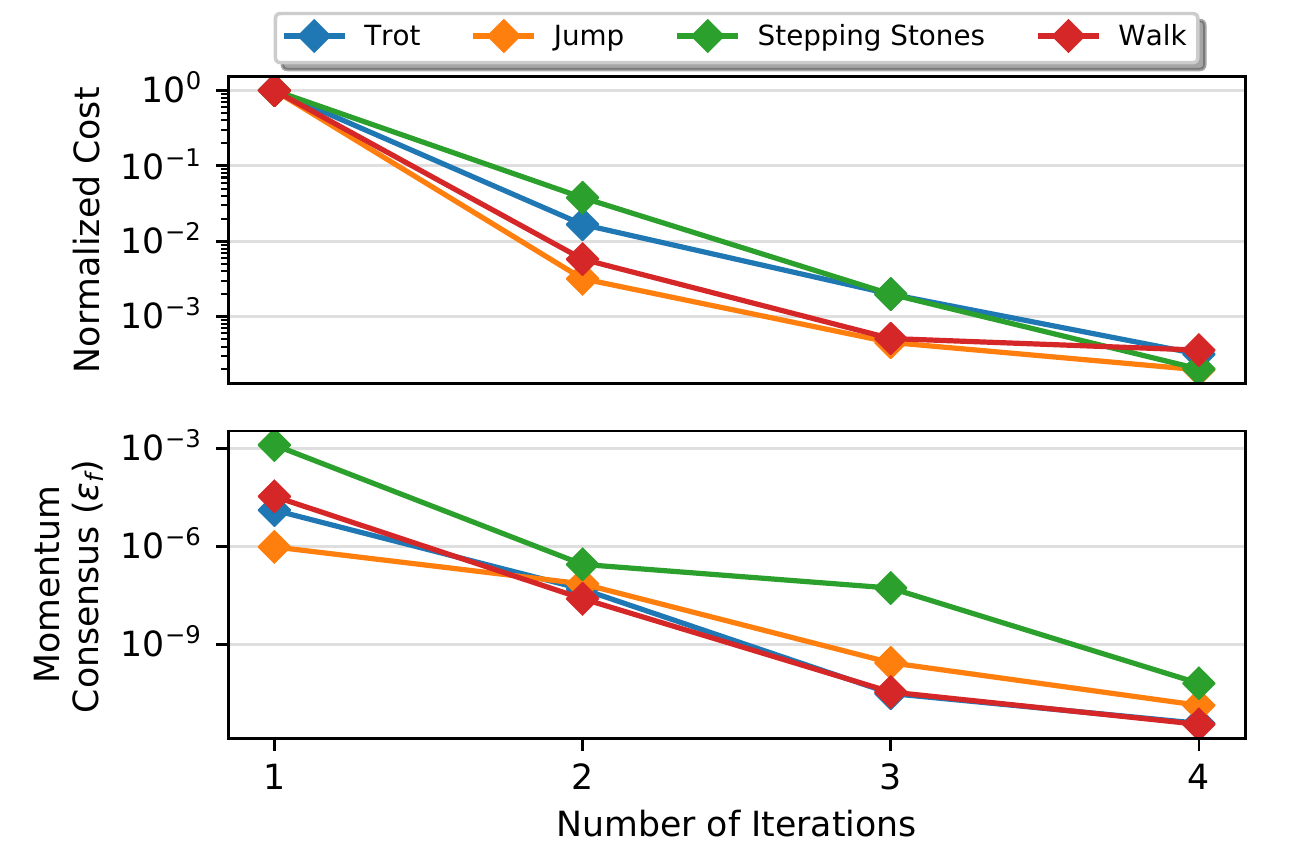}
    \caption{In the top graph, we show the normalized cost per iteration for a variety of motions. We can see that our algorithm tends to converge after the 3rd major iteration, although every iteration after the first is technically feasible. On the bottom, we graph our convergence criteria $\epsilon_{f}$. By the end of the second iteration, we find that generally our momentum profiles $\bm{\ell}$ find consensus to a high enough tolerance that these motions can be tracked by our WBC.}
    \label{costperit}
\end{figure}

We further tested the ability of our algorithm to generate arbitrary motions for navigating uneven terrain. We tested our algorithm on multiple staircases with the height of each staircase ranging in values from 12.5\% to 35\% of the center of mass height of the robot. For navigating both up and down stairs, our algorithm converged within two iterations which took \SI{0.827}{\second} for a horizon of $N=300$ which coincided with a 3 second horizon.

We also computed motions for non-coplanar terrain such as angled stepping stones as in \autoref{fig:intro_figure}. We noticed that as the contact angles became more aggressive, our algorithm took longer to converge, often requiring three major iterations (i.e. iterations of \autoref{Algorithm 1}) with total solve times on average \SI{2.398}{\second} for $N=300$, which is also due to the individual QPs requiring more iterations to converge. 
Finally, we also tested the ability of our algorithm to generate highly dynamic motions with flight phases. We were able to generate and track trajectories for jumping in place as well as jumping forward and rotationally (around the $z$-axis). We note the importance of regulating angular momentum in directional jumps. In particular, although a trajectory may be feasible, regulation of angular momentum plays a large role such that the robot does not flip in the air. We found that even during fairly aggressive jumps, we were able to generate motions that generated minimal angular momentum and could thus be successfully tracked.

\Cref{fig:multicontact_traj} shows the results of a non-gaited multi-contact motion that requires navigating uneven terrain with a jump off the top step. Despite the multiple contact switches and dynamic motion, our WBC was able to track such a motion successfully. All the resulting motions simulated in pyBullet are shown in the accompanying video.

\subsection{Evaluation of convergence}
We plot the cost per iteration for 50 motions from the categories above (various legged gaits, uneven terrain, and jumping) with a range of $L_{i}$ from $[100 : 1,000,000]$ and $\alpha=100$ in \autoref{costperit}. As we can see, both our cost and momentum profiles tend to converge and find a lower bound after three to four iterations. 

Figure \ref{costperit} also shows how our momentum consensus parameter, $\bm{\epsilon}_{f}$ changes per iteration. For the problems of interest, we experimentally found that a good value for our convergence parameter $\epsilon_{f}$ was $10^{-7}$. Using the above values for $L_{i}$ and $\alpha$, for every motion we generated, we were able to converge within a maximum of three iterations. 

\subsection{Speed and scaling}
\begin{figure}
    \center
    \includegraphics[width=\linewidth]{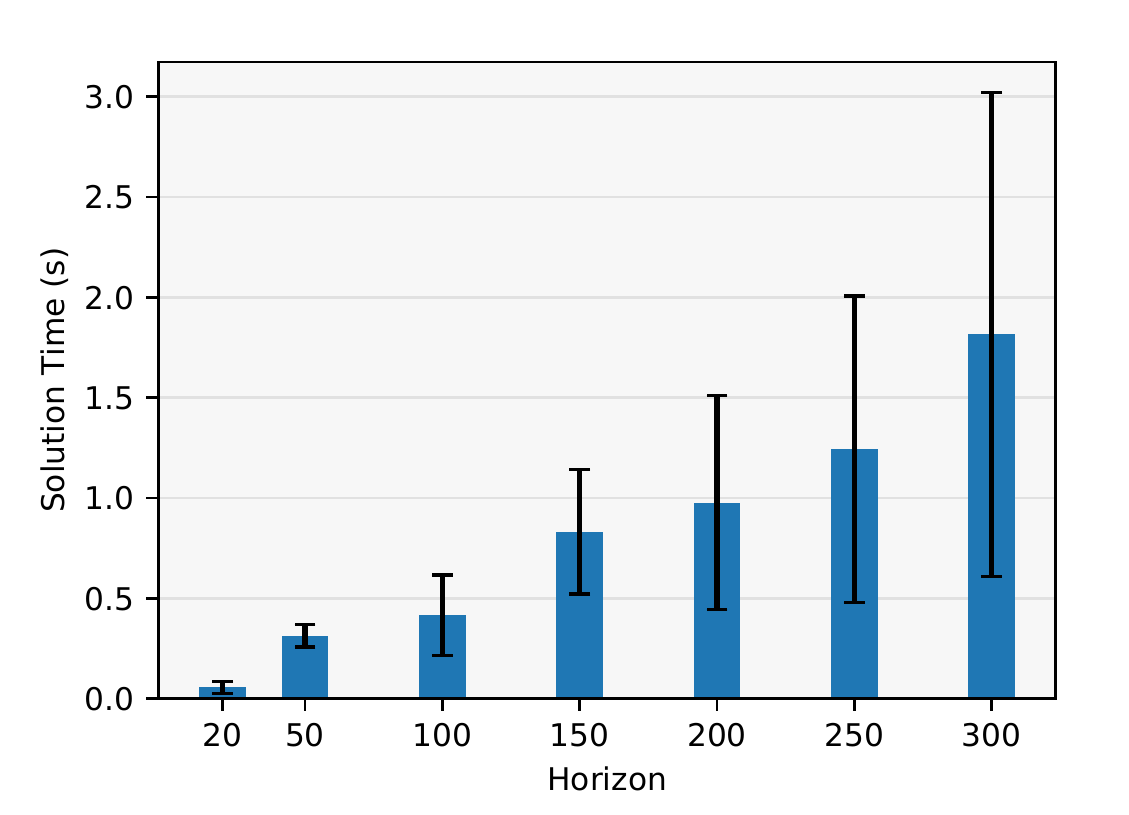}
    \caption{The total solution time for our algorithm for a variety of 50 motions. We note a roughly linear increase in time with horizon for the problem sizes we are interested in due to the sparsity of our problem. Our solution time contains only the solve time of each QP and does not include the setup time of each QP.}
    \vspace{-8pt}
    \label{fig:solve_time}
    \vspace{-6pt}
\end{figure}

We plot the total solution time of the algorithm in \autoref{fig:solve_time}. The solution times plotted are the total solve times of each QP and do not include the time to modify the appropriate QP from their previous solutions as these are trivial. We also do not include the initial setup time of the solver for each problem as this can be set beforehand with the predefined, sparse structure before running the actual optimization and, thus, only has to be run once for each QP. In general we found that for simple motions such as walking and trotting, our algorithm tended to converge within 2 overall iterations which on average took \SI{0.7972}{\second} for $N=300$. For navigation of non-coplanar terrain we saw that an extra iteration was often required; we believe this was due to the the tightness of the constraint set of our trajectories (e.g. due to the rotated friction cones). For $N=300$, navigation of non-coplanar terrain averaged \SI{2.647}{\second}. 

The algorithm spends the majority of the time of each iteration on the Force-QP. For motions with a horizon of $N=300$, the algorithm spends on average \SI{97.8}{\percent} of the total solution time on the Force-QP and \SI{2.2}{\percent} on the Contact-QP despite their similar size. This is due to the fewer number of constraints in the Contact-QP.

Finally, we explore the scalability of our method for different lengths of time horizons $N$ and show the result in \autoref{fig:solve_time}. We observed an approximately linear-time complexity in solve time with respect to horizon length. This is due to the banded, sparse nature of our problem which can be efficiently exploited by sparse QP solvers.

\begin{table*}[ht!]
\centering
\begin{tabular} {|| c | c | c | c | c ||}
    \hline
    Solver Method & Average Normalized Cost & Solve Time ($N=300$) & Change In Solve Time & Solver Success Rate \\
    \hline
    Block Coordinate Descent & $\mathbf{1.00}$ & $\mathbf{1.77 \pm 1.23 \si{\second}}$ & -- & $\mathbf{100\%}$ \\
    \hline
    Sequential Convex Relaxations & $1.08$ & $7.34 \pm 3.34 \si{\second}$ & {4.15x} slower & $\mathbf{100\%}$ \\
    \hline
    Interior Point Method (IPOPT) & $1.34$  & $8.12 \pm 2.32 \si{\second}$ & {4.59x} slower & $92\%$ \\
    \hline
\end{tabular}
\vspace{2pt}
\caption{Comparison of methods for solving the centroidal dynamics optimization problem.}
\label{table:comparison}
\vspace{-22pt}
\end{table*}
\subsection{Comparison to other solution methods}
In this section, we compare the solutions found between our block coordinate descent method proposed above, the soft constraint sequential convex relaxation (SCR) method outlined in \cite{SCR}, and the primal-dual interior point method used in IPOPT which is often used in locomotion research \cite{TOWR,RPC}. For the following comparisons, we configured IPOPT with the sparse symmetric linear solver MA57. Although \cite{SCR} provided multiple heuristics, we chose to use the soft constraints as in our experience these performed better than the trust region method. We provide this comparison as a means to quantify the solutions found by our method to against approaches where the solutions are associated with guaranteed local minima and not as a means to compare the quality of the solutions.

We first compare the final cost obtained for a variety of different multi-contact locomotion problems including flat ground motions, uneven terrain, and jumping motions. We note that the final cost calculated using the block coordinate descent method uses the original cost function, without the proximal term. Specifically, we use the solution generated by \autoref{Algorithm 1}, and then plug this back into the original cost function, \autoref{traj_opt_cost}, to evaluate the cost. We found that our method generally finds lower costs than both alternative approaches despite not converging to a local minimum. \par

Next, we compare the solution time of each method. Once again, we evaluate our speed for a variety of different motions. Our method readily outperformed both and is on average more than four times faster than than both the SCR method and IPOPT for trajectories with horizons of $N=300$. 

The results of both comparisons can be found in \autoref{table:comparison}. We note that the discrepancy in solve times between our reported values and those reported in \cite{SCR} are likely due to the types of motions we generate as well as the default tolerances and settings of the open source implementation. We would also like to point out that for the problem sizes we are interested in, the authors of \cite{SCR} noted a roughly linear time complexity in the problem horizon solve time due to the sparse nature of our problem (similar to our BCD implementation). Due to the use of a sparse solver in our IPOPT implementation we expect to see a similar trend.

\begin{figure}
    \center
    \includegraphics[width=\linewidth]{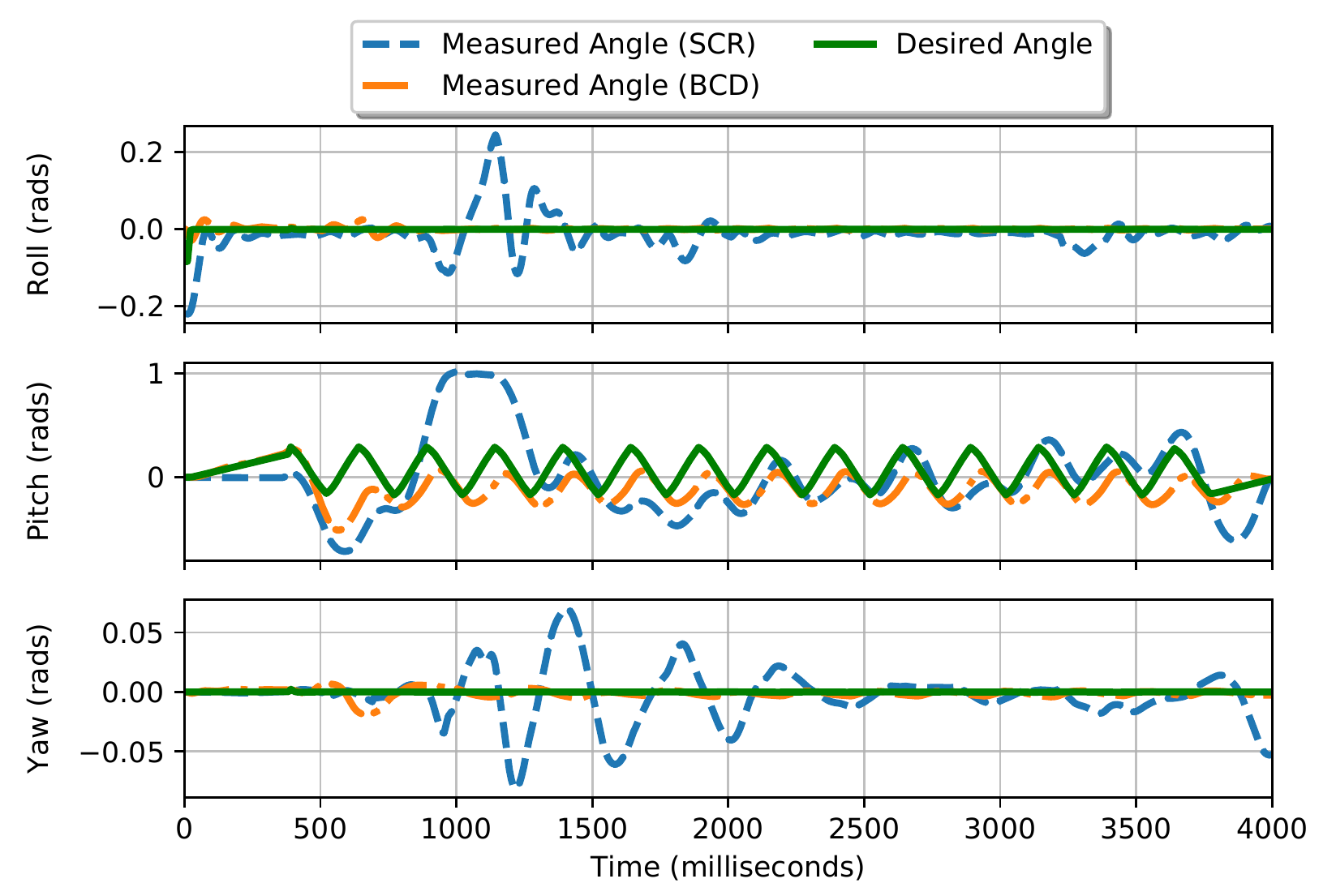}
    \caption{The tracking ability of a planned bounding gait between the SCR method and our BCD method. In order to generate this motion, we gave each method a reference angular momentum corresponding to a maximum pitch of \SI{\pm15}{\degree}. The BCD was able to create a motion that could be tracked by our WBC (yellow), however, the SCR method was unable to do so (blue). Specifically, we note the large error starting at \SI{1000}{\milli\second}. The RMSE (Root Mean Square Error) of the SCR pitch tracking is $5.84\cdot10^{-3}$ whereas the RMSE for the BCD approach is $3.33\cdot10^{-3}$ (\SI{43}{\percent} lower). We note that the desired angle was converted into a desired angular momentum using the matrix logarithm.}
    \vspace{-6pt}
    \label{fig:bounding_comparison}
    \vspace{-8pt}
\end{figure}

Of particular note is the ability of our solver to find motions that require finer tracking of angular momentum such as bounding. Unlike the SCR method, our method was able to create trajectories for such motions that could be tracked by our WBC reasonably well in simulation. We plot a comparison of the results of tracking an open-loop bounding trajectory in \autoref{fig:bounding_comparison}.
From a numerical optimization standpoint, we found that, compared to the SCR method, BCD solutions were often satisfied to much higher numerical tolerances, especially for dynamic motions such as jumping and bounding. Finally, we note that IPOPT failed to find solutions for \SI{8}{\percent} of the given multi-contact motions whereas both the BCD and SCR methods were able to find solutions for all the given scenarios.



\section{Discussion and Conclusion} \label{sec:discussion}
The experiments above indicate that despite the lack of guarantees of convergence to a critical point, our algorithm is still able to generate high quality multi-contact profiles that can be tracked by our robot. Although guarantees of global convergence are nice to have, in practice, the robotics community uses tools such as trajectory optimization as motion generators and the cost functions used are often arbitrary.

Due to the structured, sparse nature of our problem, we also see a significant speedup that can be exploited by off-the-shelf quadratic solvers. OSQP, for example, employs factorization caching which is utilized when re-solving the linear system of equations as well as polishing, a feature to predict the number of active constraints. We believe that as quadratic programming solvers continue to mature, we may be able to further exploit these types of features to further speed up our solve times.



%
We presented a novel method for solving the non-convex centroidal dynamics optimization problem. Rather than relying on off-the-shelf nonlinear solvers which have no convergence guarantees or complicated relaxation methods, we showed that by applying BCD and using a standard quadratic programming solver, we are able to efficiently find reasonable solutions for the non-convex problem. Compared to the state of the art, we are also able to solve the problem multiple times faster and can often generate and track trajectories that require finer tracking of angular momentum. 

We believe our algorithm is well-suited for model predictive control or variable horizon control. In particular, we can exploit our algorithm's capability of finding feasible trajectories at every iteration to reduce the number of cycles required even further. We are also able to warm-start solutions from previous solutions which may further decrease our solve time. Due to the use of interior point methods, this ability cannot be exploited by the relaxation methods proposed in \cite{SCR} as well as frameworks built upon IPOPT. Additionally, these properties also enable us to efficiently combine our method with pre-computed libraries or data-driven techniques \cite{viereck2020learning,trajectory_of_libraries,merkt2018leveraging}.
Finally, we believe BCD is a versatile approach for trajectory optimization in robotics due to its simplicity of implementation. In the future, we plan to validate our trajectories using hardware experiments and extend our method to real-time control.




\section*{Appendix} \label{sec:appendix}


\begin{table}[H]
    \centering
    \begin{tabular}{|c|c|}
        \hline
        \multicolumn{2}{|c|}{\bf{OSQP Solver Settings}} \\
        \hline
         Absolute Tolerance, $\epsilon_{abs}$ &  $1e-7$\\
         \hline
         Relative Tolerance, $\epsilon_{rel}$ & $1e-7$\\
         \hline
         Primal Infeasibility Tolerance, $\epsilon_{prim \, inf}$ & $1e-6$\\
         \hline
         Dual Infeasibility Tolerance, $\epsilon_{dual \, inf}$ & $1e-6$\\
         \hline
         Polish & True \\
         \hline
         Scaled Termination & True \\
         \hline
         Adaptive Rho & True \\
         \hline
         Check Termination & 50 \\
         \hline
    \end{tabular}
    \vspace{10pt}
    \caption{OSQP Solver settings for force and contact QP}
    \label{table:1}
    \vspace{-10pt}
\end{table}

\section*{Acknowledgments}

The authors would like to thank Julian Viereck for his assistance with the kino-dynamic planner.


\bibliographystyle{IEEEtran}
\bibliography{IEEEfull,references}

\begin{thebibliography}{10}
\providecommand{\url}[1]{#1}
\csname url@rmstyle\endcsname
\providecommand{\newblock}{\relax}
\providecommand{\bibinfo}[2]{#2}
\providecommand\BIBentrySTDinterwordspacing{\spaceskip=0pt\relax}
\providecommand\BIBentryALTinterwordstretchfactor{4}
\providecommand\BIBentryALTinterwordspacing{\spaceskip=\fontdimen2\font plus
\BIBentryALTinterwordstretchfactor\fontdimen3\font minus
  \fontdimen4\font\relax}
\providecommand\BIBforeignlanguage[2]{{%
\expandafter\ifx\csname l@#1\endcsname\relax
\typeout{** WARNING: IEEEtran.bst: No hyphenation pattern has been}%
\typeout{** loaded for the language `#1'. Using the pattern for}%
\typeout{** the default language instead.}%
\else
\language=\csname l@#1\endcsname
\fi
#2}}

\bibitem{template_full}
R.~Full and D.~Koditschek, ``Templates and anchors: neuromechanical hypotheses
  of legged locomotion on land,'' \emph{Journal of Experimental Biology}, vol.
  202, no.~23, pp. 3325--3332, 1999.

\bibitem{Kajita}
S.~{Kajita}, F.~{Kanehiro}, K.~{Kaneko}, K.~{Fujiwara}, K.~{Harada},
  K.~{Yokoi}, and H.~{Hirukawa}, ``Biped walking pattern generation by using
  preview control of zero-moment point,'' in \emph{Proc. IEEE Int. Conf. Rob.
  Autom. (ICRA)}, vol.~2, 2003, pp. 1620--1626 vol.2.

\bibitem{herdt}
A.~Herdt, H.~Diedam, P.-B. Wieber, D.~Dimitrov, K.~Mombaur, and M.~Diehl,
  ``Online walking motion generation with automatic footstep placement,''
  \emph{Advanced Robotics}, vol.~24, no. 5-6, pp. 719--737, 2010.

\bibitem{englesberger}
J.~{Englsberger}, C.~{Ott}, and A.~{Albu-Schäffer}, ``Three-dimensional
  bipedal walking control based on divergent component of motion,''
  \emph{{IEEE} Transactions on Robotics}, vol.~31, no.~2, pp. 355--368, 2015.

\bibitem{hopkins}
M.~A. {Hopkins}, D.~W. {Hong}, and A.~{Leonessa}, ``Humanoid locomotion on
  uneven terrain using the time-varying divergent component of motion,'' in
  \emph{Proc. IEEE-RAS Int. Conf. Hum. Rob. (Humanoids)}, 2014, pp. 266--272.

\bibitem{orin}
D.~Orin, A.~Goswami, and S.-H. Lee, ``Centroidal dynamics of a humanoid
  robot,'' \emph{Autonomous Robots}, vol.~35, 10 2013.

\bibitem{dai_fullbody}
H.~{Dai}, A.~{Valenzuela}, and R.~{Tedrake}, ``Whole-body motion planning with
  centroidal dynamics and full kinematics,'' in \emph{Proc. IEEE-RAS Int. Conf.
  Hum. Rob. (Humanoids)}, 2014, pp. 295--302.

\bibitem{kino-dyn}
A.~Herzog, S.~Schaal, and L.~Righetti, ``Structured contact force optimization
  for kino-dynamic motion generation,'' in \emph{Proc. IEEE/RSJ Int. Conf.
  Intell. Rob. Sys. (IROS)}, 10 2016, pp. 2703--2710.

\bibitem{carpentier}
J.~{Carpentier} and N.~{Mansard}, ``Multicontact locomotion of legged robots,''
  \emph{{IEEE} Transactions on Robotics}, vol.~34, no.~6, pp. 1441--1460, 2018.

\bibitem{SCR}
B.~{Ponton}, M.~{Khadiv}, A.~{Meduri}, and L.~{Righetti}, ``{Efficient
  Multi-Contact Pattern Generation with Sequential Convex Approximations of the
  Centroidal Dynamics},'' \emph{IEEE Transactions on Robotics}, pp. 1--19,
  2021.

\bibitem{dai}
H.~{Dai} and R.~{Tedrake}, ``Planning robust walking motion on uneven terrain
  via convex optimization,'' in \emph{Proc. IEEE-RAS Int. Conf. Hum. Rob.
  (Humanoids)}, 2016, pp. 579--586.

\bibitem{ecos}
A.~{Domahidi}, E.~{Chu}, and S.~{Boyd}, ``Ecos: An socp solver for embedded
  systems,'' in \emph{European Control Conference (ECC)}, 2013, pp. 3071--3076.

\bibitem{multi_convex_boyd}
X.~{Shen}, S.~{Diamond}, M.~{Udell}, Y.~{Gu}, and S.~{Boyd}, ``Disciplined
  multi-convex programming,'' in \emph{2017 29th Chinese Control And Decision
  Conference (CCDC)}, 2017, pp. 895--900.

\bibitem{ipopt}
A.~Wächter and L.~Biegler, ``On the implementation of an interior-point filter
  line-search algorithm for large-scale nonlinear programming,''
  \emph{Mathematical programming}, vol. 106, pp. 25--57, 03 2006.

\bibitem{bcd}
Y.~Xu and W.~Yin, ``A block coordinate descent method for regularized
  multiconvex optimization with applications to nonnegative tensor
  factorization and completion,'' \emph{SIAM Journal on Imaging Sciences},
  vol.~6, no.~3, pp. 1758--1789, 2013.

\bibitem{wieber2006holonomy}
P.-B. Wieber, ``Holonomy and nonholonomy in the dynamics of articulated
  motion,'' in \emph{Fast motions in biomechanics and robotics}.\hskip 1em plus
  0.5em minus 0.4em\relax Springer, 2006, pp. 411--425.

\bibitem{momentum_id}
A.~Herzog, N.~Rotella, S.~Mason, F.~Grimminger, S.~Schaal, and L.~Righetti,
  ``Momentum control with hierarchical inverse dynamics on a torque-controlled
  humanoid,'' \emph{Autonomous Robots}, vol.~40, 10 2014.

\bibitem{deits}
R.~{Deits} and R.~{Tedrake}, ``Footstep planning on uneven terrain with
  mixed-integer convex optimization,'' in \emph{Proc. IEEE-RAS Int. Conf. Hum.
  Rob. (Humanoids)}, 2014, pp. 279--286.

\bibitem{sl1m}
S.~Tonneau, D.~Song, P.~Fernbach, N.~Mansard, M.~Taïx, and A.~Del~Prete,
  ``{SL1M: Sparse L1-norm Minimization for contact planning on uneven
  terrain},'' in \emph{Proc. IEEE Int. Conf. Rob. Autom. (ICRA)}, 05 2020, pp.
  6604--6610.

\bibitem{lin2018humanoid}
Y.-C. Lin and D.~Berenson, ``Humanoid navigation planning in large unstructured
  environments using traversability-based segmentation,'' in \emph{2018
  IEEE/RSJ International Conference on Intelligent Robots and Systems
  (IROS)}.\hskip 1em plus 0.5em minus 0.4em\relax IEEE, 2018, pp. 7375--7382.

\bibitem{gauss-seidel}
G.~H. Golub and C.~F. Van~Loan, \emph{Matrix Computations (3rd Ed.)}.\hskip 1em
  plus 0.5em minus 0.4em\relax USA: Johns Hopkins University Press, 1996.

\bibitem{kino-dyn-github}
B.~Ponton, ``Kino-dynamic trajectory optimization for multiped robots,''
  \url{https://github.com/machines-in-motion/kino_dynamic_opt}, accessed:
  2020-11-30.

\bibitem{solo}
F.~{Grimminger}, A.~{Meduri}, M.~{Khadiv}, J.~{Viereck}, M.~{Wüthrich},
  M.~{Naveau}, V.~{Berenz}, S.~{Heim}, F.~{Widmaier}, T.~{Flayols}, J.~{Fiene},
  A.~{Badri-Spröwitz}, and L.~{Righetti}, ``An open torque-controlled modular
  robot architecture for legged locomotion research,'' \emph{{IEEE} Robotics
  and Automation Letters}, vol.~5, no.~2, pp. 3650--3657, 2020.

\bibitem{pybullet}
E.~Coumans and Y.~Bai, ``{PyBullet, a Python module for physics simulation for
  games, robotics and machine learning},'' \url{http://pybullet.org},
  2016--2019.

\bibitem{osqp}
B.~Stellato, G.~Banjac, P.~Goulart, A.~Bemporad, and S.~Boyd, ``{OSQP}: an
  operator splitting solver for quadratic programs,'' \emph{Mathematical
  Programming Computation}, vol.~12, no.~4, pp. 637--672, 2020.

\bibitem{TOWR}
A.~W. {Winkler}, C.~D. {Bellicoso}, M.~{Hutter}, and J.~{Buchli}, ``Gait and
  trajectory optimization for legged systems through phase-based end-effector
  parameterization,'' \emph{{IEEE} Robotics and Automation Letters}, vol.~3,
  no.~3, pp. 1560--1567, 2018.

\bibitem{RPC}
G.~{Bledt} and S.~{Kim}, ``Implementing regularized predictive control for
  simultaneous real-time footstep and ground reaction force optimization,'' in
  \emph{Proc. IEEE/RSJ Int. Conf. Intell. Rob. Sys. (IROS)}, 2019, pp.
  6316--6323.

\bibitem{viereck2020learning}
J.~Viereck and L.~Righetti, ``Learning a centroidal motion planner for legged
  locomotion,'' 2020.

\bibitem{trajectory_of_libraries}
M.~{Stolle} and C.~G. {Atkeson}, ``Policies based on trajectory libraries,'' in
  \emph{Proc. IEEE Int. Conf. Rob. Autom. (ICRA)}, 2006, pp. 3344--3349.

\bibitem{merkt2018leveraging}
W.~Merkt, V.~Ivan, and S.~Vijayakumar, ``Leveraging precomputation with problem
  encoding for warm-starting trajectory optimization in complex environments,''
  in \emph{Proc. IEEE/RSJ Int. Conf. Intell. Rob. Sys. (IROS)}, Oct 2018, pp.
  5877--5884.

\end{thebibliography}

\end{document}